\documentclass[10pt,twocolumn,letterpaper]{article}

\usepackage{iccv}
\usepackage{times}
\usepackage{epsfig}
\usepackage{graphicx}
\usepackage{amsmath}
\usepackage{amssymb}

\usepackage{multirow} 
\usepackage{booktabs}
\usepackage{subfig}
\usepackage[breaklinks=true,bookmarks=false]{hyperref}

\iccvfinalcopy 

\def\iccvPaperID{***} 

\ificcvfinal\fi

\begin{document}

\title{Self-NeRF: A Self-Training Pipeline for Few-Shot Neural Radiance Fields}

\author{Jiayang Bai\\
Nanjing University\\
{\tt\small jybai@smail.nju.edu.cn}
\and
Letian Huang\\
Central South University\\
{\tt\small lthuangg@gmail.com}
\and
Wen Gong\\
Nanjing University\\
{\tt\small  lesliewinnie@163.com}
\and
Jie Guo\\
Nanjing University\\
{\tt\small  jieguo@nju.edu.cn}
\and
Yanwen Guo\\
Nanjing University\\
{\tt\small  ywguo@nju.edu.cn}
}

\ificcvfinal\thispagestyle{empty}\fi
\twocolumn[{
\renewcommand\twocolumn[1][]{#1}
\maketitle
\begin{center}
    \captionsetup{type=figure}
    \renewcommand\tabcolsep{1.0pt}
  \begin{tabular}{ccccc}
  \includegraphics[width=0.2\linewidth, trim={0px, 0px, 0px, 0px}, clip]{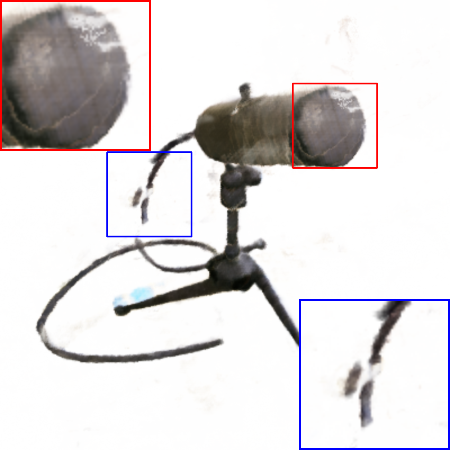}
  &\includegraphics[width=0.2\linewidth, trim={0px, 0px, 0px, 0px}, clip]{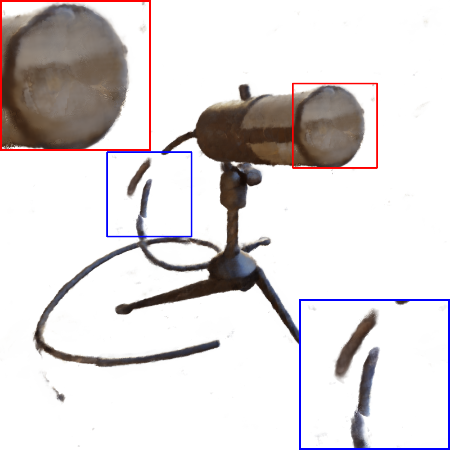}
  &\includegraphics[width=0.2\linewidth, trim={0px, 0px, 0px, 0px}, clip]{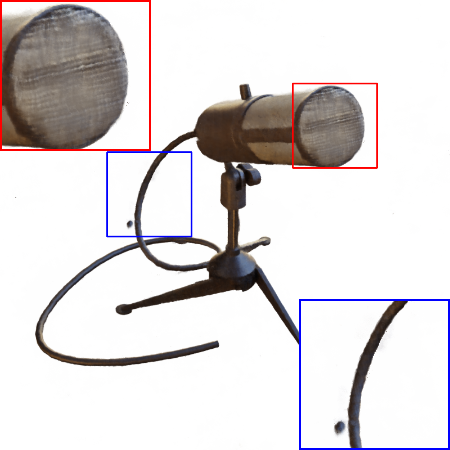}
  &\includegraphics[width=0.2\linewidth, trim={0px, 0px, 0px, 0px}, clip]{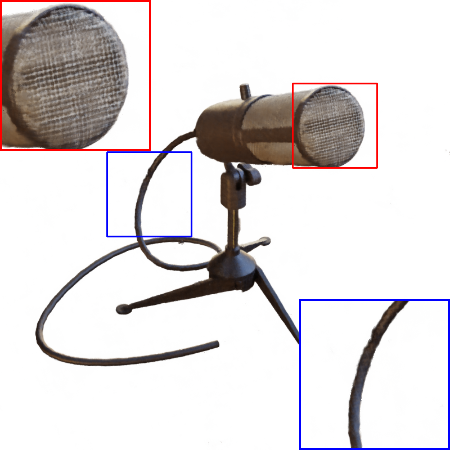}
  &\includegraphics[width=0.2\linewidth, trim={0px, 0px, 0px, 0px}, clip]{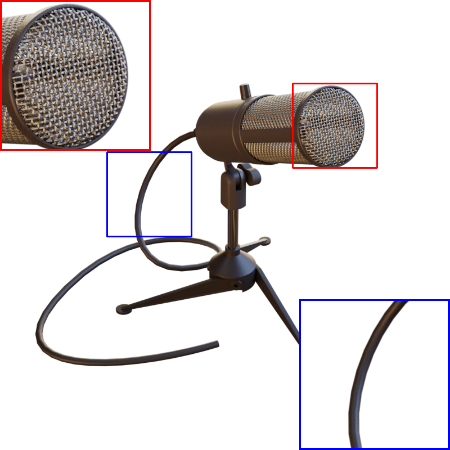}\\
  DietNeRF~\cite{Jain2021PuttingDiet} & InfoNeRF~\cite{InfoNeRF} &Ours (2 iterations)&Ours (6 iterations) & Ground truth\\
  \end{tabular}
  \captionof{figure}{Qualitative comparison of different methods on the mic scene in 4-view settings.  Compared with other methods, Self-NeRF yields a more realistic rendering with superior details. Note that Self-NeRF produces sharp details and reduces artifacts through iterative training.}
  \label{fig:teaser}
\end{center}
}]
\begin{abstract}
Recently, Neural Radiance Fields (NeRF) have emerged as a potent method for synthesizing novel views from a dense set of images. Despite its impressive performance, NeRF is plagued by its necessity for numerous calibrated views and its accuracy diminishes significantly in a few-shot setting. To address this challenge, we propose Self-NeRF, a self-evolved NeRF that iteratively refines the radiance fields with very few number of input views, without incorporating additional priors. Basically, we train our model under the supervision of reference and unseen views simultaneously in an iterative procedure. In each iteration, we label unseen views with the predicted colors or warped pixels generated by the model from the preceding iteration. However, these expanded pseudo-views are afflicted by imprecision in color and warping artifacts, which degrades the performance of NeRF. To alleviate this issue, we construct an uncertainty-aware NeRF with specialized embeddings. Some techniques such as cone entropy regularization are further utilized to leverage the pseudo-views in the most efficient manner. Through experiments under various settings, we verified that our Self-NeRF is robust to input with uncertainty and surpasses existing methods when trained on limited training data. 
\end{abstract}

\section{Introduction}
Synthesizing novel camera views given a set of known views is an important task in computer vision and a prerequisite to many AR and VR applications. Classic techniques have addressed this problem using structure-from-motion~\cite{Harltey2003MultipleVG} or light fields~\cite{1996lightfield}.
Recently, Neural Radiance Fields (NeRF)~\cite{Mildenhall2020NeRFRS} have gained tremendous popularity due to the impressive results in photo-realistic rendering. This approach trains learning-based models implicitly embedded within a 3D geometric context and reconstructs observed images using neural rendering techniques. Albeit effective, the performance of NeRF is highly influenced by the quality and the number of training views. When the known views are limited, NeRF collapses to trivial solutions~\cite{Pan2022ActiveNeRFLW} ( \emph{e.g.}, producing zero density for the unobserved regions) and has the risk of overfitting to seen views.
To make this challenging problem tractable, previous works attempt to incorporate some additional priors, such as a semantic feature~\cite{Jain2021PuttingDiet}, ground truth depth~\cite{Deng2021DepthsupervisedNF} or normalizing flow~\cite{Niemeyer2021RegNeRFRN}. Although these models yield adequate rendering performance, these additional priors are not always valid. Kim \emph{et al.}~\cite{InfoNeRF} propose a prior-free approach that introduces ray entropy minimization and ray information gain reduction for each ray to alleviate the reconstruction inconsistency and overfitting issue. However, entropy regularization imposes sparsity on the estimated scene, resulting in artifacts and flaws in unseen viewpoints. 

In this paper, we propose Self-NeRF to solve the few-shot novel view synthesis task without additional priors. 
Our key point is to design a self-training framework, in which we jointly learn from seen views and a large number of auxiliary unseen views.
Self-training~\cite{survey} is a classic method for semi-supervised learning. It aims to learn from unlabeled data by iteratively imputing the labels for samples predicted with high confidence. 
In the novel view synthesis task, the labeled data refers to seen views while the unlabeled data refers to unseen views.
Inspired by self-training, our self-training backbone leverages confident predictions in the previous iteration to produce pseudo-views for unseen views. Our pseudo-views can be categorized into two types: the warped pseudo-views generated through forward warping and the predicted pseudo-views which are the outputs of the previous iteration.
The former provide local texture guidance, while the latter help improve the perceptual capability of the global structures. In other words, pseudo-views add more information to guide model training. 

However, these pseudo-views still contain uncertain regions with inaccurate colors due to warping artifacts or occlusion. Assuming that all input pixel colors are reliable, NeRF will faithfully learn these uncertain pixels in  pseudo-views, which results in reconstruction inconsistency across multiple views and degenerate solutions.
To avoid this degeneracy, we propose an uncertainty-aware NeRF that autonomously learns a field of uncertainty from pseudo-views. Based on the output of the uncertainty field, we alleviate the impact of uncertain pixels. Specifically, we introduce our specialized warping embeddings and uncertainty embeddings to model image-dependent uncertain colors.
Furthermore, we leverage cone tracing technique and a cone entropy regularization within a conical frustum to represent fine details. The cone entropy regularization imposes the model to compact representation in the unobserved views instead of collapsing to a trivial solution.
Our experiments have proved that the proposed Self-NeRF shows state-of-the-art performance on few-shot novel view synthesis as shown in Fig.~\ref{fig:teaser}. 

In summary, the main contributions of our paper are:
\begin{itemize}
    \item We propose Self-NeRF, a novel iterative training scheme for synthesizing novel views from few-shot images without additional priors.
    \item We prove the convergence of our iterative Self-NeRF through theoretical deduction and experiments.
    \item We introduce a practical implementation of Self-NeRF, which leverages an uncertainty-aware NeRF, specialized embeddings and a cone entropy regularization to avoid degeneration due to pseudo-views.
\end{itemize}
\section{Related Work}
\subsection{Novel view synthesis}
Given a dense sampling of views,  earlier works use view interpolation~\cite{Chen1993ViewIF} and light fields ~\cite{1996lightfield, Gortler1996TheL} to reconstruct novel views. To better represent the 3D scene, some works utilize proxy geometry~\cite{Debevec1996ModelingAR} and explicit representation such as layered representations ~\cite{Shade1998LayeredDI, Shih20203DPU}, voxel~\cite{Sitzmann2018DeepVoxelsLP}, mesh~\cite{Jack2018LearningFD} and point cloud~\cite{Qi2016PointNetDL, Wiles2019SynSinEV}.
Recently, a plethora of learning-based methods~\cite{Flynn2019DeepViewVS, Flynn2015DeepSL, Kalantari2016LearningbasedVS, Lin2021Deep3M, Zhang2018TheUE} has received growing attention. 
Simultaneously, there is another line of work that uses volumetric representations to address the task of photo-realistic view synthesis. Neural radiance fields (NeRF)~\cite{Mildenhall2020NeRFRS} 
employ an implicit neural representation of a 3D scene and use volumetric rendering to generate photo-realistically unseen views. 
Mip-NeRF~\cite{mipnerf} follows the step of NeRF and reasons about volumetric frustums along a cone for anti-aliasing. Mip-NeRF 360~\cite{Barron2021MipNeRF3U} further extends it to model unbounded scenes with a non-linear scene parameterization, online distillation, and a novel distortion-based regularizer.
Furthermore, recent works have made tremendous efforts to improve the rendering speed~\cite{Mller2022InstantNG, Reiser2021KiloNeRFSU, Sun2021DirectVG, Yu2021PlenoxelsRF}, artistic effects~\cite{Wang2021CLIPNeRFTD, Fan2022UnifiedIN, Zhang2022ARFAR}, and generalization ability ~\cite{Liu2021NeuralRF, Wang2021IBRNetLM} of NeRF. Some works~\cite{NeRFW, Wu2022D2NeRFSD, Zhang2022FDNeRFFD} also adapt NeRF for dynamic scenes.
\begin{figure*}[htp]
  \centering
   \includegraphics[width=1.0\linewidth]{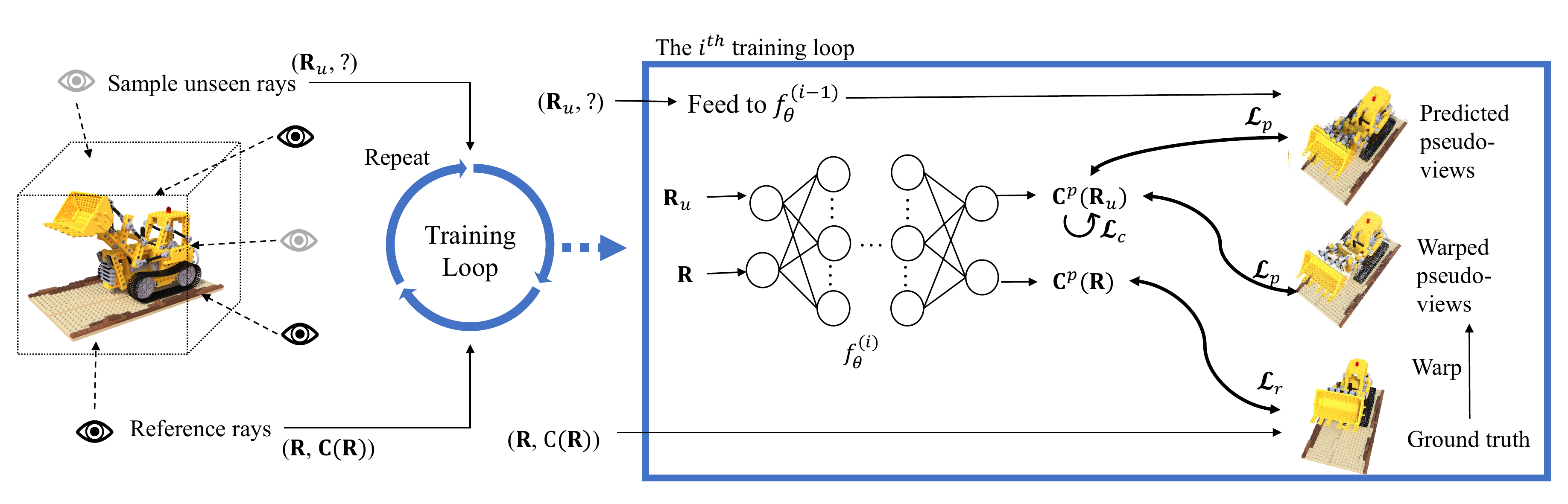}
   \caption{The iterative pipeline for Self-NeRF. In $i^{th}$ iteration, we gather predicted pseudo-views synthesized by the model in the previous iteration for unseen views. The warped pseudo-views are collected through warping the seen views to unseen views. Then we train the uncertainty-aware NeRF $f^{i}_{\theta}$ in a supervised way with seen views and pseudo-views simultaneously.}
   \label{fig:pipeline}
\end{figure*}
\subsection{Few-shot view synthesis}
The requirement of numerous calibrated images is a key limitation of NeRF. 
Some works~\cite{Deng2021DepthsupervisedNF, Roessle2021DenseDP} attempt to decrease the data-hungriness by using the depth priors.
Without the depth priors, MetaNeRF~\cite{Li2020MetaNERNE} uses data-driven priors recovered from a domain of training scenes. PixelNeRF~\cite{Yu2020pixelNeRFNR} and Peng \emph{et al.}~\cite{Peng2020NeuralBI} utilize the implicit spatial information in the local CNN features to construct the radiance fields. MVSNeRF~\cite{Chen2021MVSNeRFFG} and IBRNet~\cite{Wang2021IBRNetLM} employ earlier multi-view stereo methods to produce a multi-view feature volume. 
DietNeRF~\cite{Jain2021PuttingDiet} resorts to the pretrained CLIP-ViT~\cite{Radford2021LearningTV} and adapts its projected image embeddings as features to add global semantic information for novel views. 
These methods heavily rely on external supervisory signals such as depth information or additional pretrained encoders to synthesize novel views. 
Recently, some studies~\cite{Bortolon2022DataAug, Ahn2022PANeRFPA, sinnerf, Darmon2021ImprovingNI} have introduced the augmentation of warped views with only few-shot images to improve neural radiance fields, which provides more training constraints designed in a small patch. However, these methods achieve suboptimal performance since they ignore the uncertainty of warped views caused by the warping operations.  
Unlike the aforementioned works, some methods~\cite{InfoNeRF, Niemeyer2021RegNeRFRN, Chen2022StructNeRFNR, Fu2022GeoNeusGN} introduce a prior-free model. InfoNeRF~\cite{InfoNeRF} minimizes the ray entropy among seen and unseen poses and utilizes ray information gain reduction to alleviate reconstruction inconsistency across views. Although their strategies improve the quality of novel-view synthesis, they do not fully explore the full potential of unseen views. Artifacts such as blurring and cloud effects still exist in the synthesized image. By contrast, we fully leverage adequate pseudo-views and reduce the artifacts with our iterative training.
\section{Method}
\subsection{Preliminaries}\label{sec:background}
Neural radiance fields represent a 3D scene as a continuous implicit function $f_\theta$, which outputs emitted radiance value and volume density when given a 3D position $\mathbf{x} \in \mathbb{R}^3$ and unit viewing direction $\mathbf{d} \in \mathbb{R}^3$. 
In practice, NeRF adapts a multi-layer perceptron (MLP) model to predict the corresponding volume density $\sigma \in [0,\infty]$ and color $\mathbf{c} \in [0,1]^3$ given the queried $\mathbf{x}$ and $\mathbf{d}$ as follows:
\begin{equation}
(\sigma, \mathbf{c}) = f_\theta(\gamma(\mathbf{x}), \gamma(\mathbf{d}))
\end{equation}
where $\gamma$ is a predefined positional encoding.

To render the RGB color at the target pixel, NeRF samples points along the ray $\bold{R}$ and integrates colors and densities based on the volume rendering. The ray $\mathbf{r}(t) = \mathbf{o} + t\mathbf{d}$ is emitted from the camera’s center $\mathbf{o}$ along the direction $\mathbf{d}$.  We compute the 3D position $\mathbf{x}_k=\mathbf{r}(t_k)$ for each sample point $t_k$. Thus the rendered color can be formulated as:
\begin{equation}\label{eq:static}
\mathbf{C}^{r}(\mathbf{R})=\sum^{N}_{i}\exp(-\sum^{i-1}_{j}\sigma_j\delta_j)(1-\exp(\sigma_i\delta_i))\mathbf{c}_i
\end{equation}
where $N$ indicates the total number of sample points along $\bold{R}$, and $\delta_i$ represents the distance between the $i^{\text{th}}$ and $(i+1)^{\text{th}}$ points.
NeRF casts a single ray per pixel and may produce renderings that are excessively blurred or aliased. To ameliorate this issue, mip-NeRF~\cite{mipnerf} casts a cone that passes through the pixel's center. To this aim,  mip-NeRF derives integrated positional encoding (IPE), which is the integration over a volume covered by a conical frustum. Mip-NeRF reduces objectionable aliasing artifacts and significantly improves NeRF’s ability to represent fine details, while also being faster meanwhile. 

Despite its impressive performance, Niemeyer \emph{et al.}~\cite{Niemeyer2021RegNeRFRN} find that the quality of mip-NeRF’s view synthesis drops significantly with only few views. Mip-NeRF fails to generalize well to novel views at test time due to training divergence. 
In addition, blurry artifacts or floaters may appear because of the inherent ambiguity of few-shot input.

\subsection{Motivation and overview}\label{sec:problem}
For the novel view synthesis task, we treat pixels in training images as labeled data while the cast rays in novel views are considered as unlabeled data.  Consequently, this task can be solved in a semi-supervised learning method.  
In this paper, we utilize an inductive semi-supervised learning method to harness large amounts of unseen views in combination with smaller sets of seen views. Specifically, we construct a novel iterative scheme in a self-learning manner~\cite{Xie2019SelfTrainingWN}.
The typical self-training algorithm~\cite{Mukherjee2020UncertaintyawareSF, Wei2021CReSTAC, Zou2019ConfidenceRS} has three main steps: 1) train a good teacher model with labeled data, 2) use the teacher model to produce pseudo-labels on unlabeled data 3) train a student model on labeled and pseudo-labeled data simultaneously. 
In Self-NeRF, we iterate this algorithm a few times by putting back the student as a teacher to relabel the unlabeled data and training a new student. 
We denote the student model in $i^{th}$ iteration as $f_\theta^{i}$, thus the trained teacher model in $i^{th}$ iteration is $f_\theta^{(i-1)}$.
In other words, the working pipeline of Self-NeRF is to train the model $f^{i}_{\theta}$ iteratively using the seen views and pseudo-views generated by $f_\theta^{(i-1)}$. The overview of Self-NeRF is shown in Fig.~\ref{fig:pipeline}. 

We observe that the performance of Self-NeRF is highly influenced by the quality of pseudo-views and the capability of the model in the iteration. To produce realistic rendering, we carefully design our pseudo-views (Sec.~\ref{sec:label}) and propose an uncertainty-aware model (Sec.~\ref{sec:model}).
Sec.~\ref{sec:loss} further describes the inference and optimization of our model. 
In addition, we discuss the convergence of our iterative training and prove the feasibility of applying self-training on the novel view synthesis task in Sec.~\ref{sec:proof}.

\subsection{Pseudo-views in Self-NeRF}\label{sec:label}
\begin{figure}[t]
  \centering
   \includegraphics[width=1.0\linewidth]{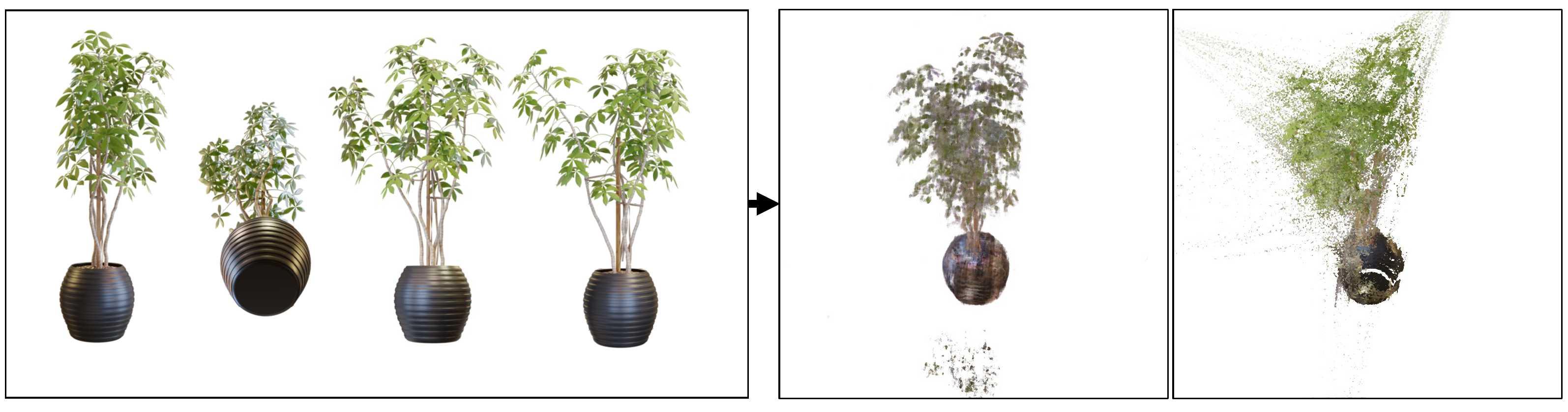}
   \caption{Given four seen views (left), we obtain the predicted pseudo-views (middle) and warped pseudo-views (right) for unseen views.}
   \label{fig:synthesized}
\end{figure}
In $i^{th}$ iteration, we first gather pseudo-views synthesized by $f^{i-1}_{\theta}$ for unseen views. As shown in Fig.~\ref{fig:synthesized}, predicted pseudo-views capture the main structure of the scene, thus adding them to training views helps improve the perceptual capability of the global structures. However, predicted pseudo-views may introduce color shifts due to training divergence, even when these pixels are visible in the training images. To alleviate this issue, we generate warped pseudo-views through the forward warping. 
In more detail, we warp seen views $\mathbf{I}_i$ to unseen views using the predicted depth map $\mathbf{D}_i$ from $f^{i-1}_{\theta}$ and get warped pseudo-views $\mathbf{I}_j$.  
For pixel $p_i\in \mathbf{I_i}$ in the seen view, the corresponding pixel $p_j$ in the unseen view is:
\begin{equation}\label{eq:warp}
    p_j=K_{j}T_{ij}(K^{-1}_{i}d_ip_i)
\end{equation}
where $d_i\in \mathbf{D}_i$ is depth of $p_i$ predicted by $f^{i-1}_{\theta}$, $K_{i}$ is the camera intrinsic matrices of $\mathbf{I}_i$ and $T_{ij}$ refers to the transform matrices from $\mathbf{I}_i$ to $\mathbf{I}_j$. Pixels in warped pseudo-views are reprojected from the seen views, thus providing local texture guidance.

\subsection{Uncertainty-aware model in Self-NeRF}\label{sec:model}
While proving global structure information and local texture guidance, pseudo-views still contain uncertain pixels for unobserved regions. Assuming that all training pixels are equally reliable, mip-NeRF tends to learn the uncertain colors. Consequently, the performance fluctuates wildly. Worse still, the low signal-to-noise ratio of pseudo-views sometimes leads to training collapse. To handle the challenges of uncertain pixels, we adapt mip-NeRF to be tolerant of uncertainty following Martin-Brualla \emph{et al.}~\cite{NeRFW}. To this aim, we model the output color $\mathbf{C}^{p}(\mathbf{R})$ as the sum of the real color $\mathbf{C}^{r}(\mathbf{R})$ and uncertain color $\mathbf{C}^{u}(\mathbf{R})$ as follows:
\begin{equation}\label{eq:sum}
    \mathbf{C}^{p}(\mathbf{R})=\mathbf{C}^{r}(\mathbf{R})+\mathbf{C}^{u}(\mathbf{R})
\end{equation}
where $\mathbf{C}^{r}(\mathbf{R})$ and $\mathbf{C}^{u}(\mathbf{R})$ are learned from the radiance field and uncertainty field respectively during training. More specifically, our model adds two specialized embeddings and a branch to emit a field of uncertainty:
\begin{equation}
(\sigma, \mathbf{c}, \sigma^{u}, \mathbf{c}^{u}, \mu^{u}) = f_\theta(\gamma(\mathbf{x}), \gamma(\mathbf{d}),\omega,\phi)
\end{equation}
where $\sigma^{u}$ and $\mathbf{c}^{u}$ are the predicted density and color of uncertainty field. $\mu^{u}$ is the uncertainty of the prediction. $\omega$ represents the learned warping embeddings that distinguish warped pseudo-views from predicted pseudo-views. Hence we assign the same $\omega$ to warped pseudo-views and seen views, explicitly implying that warped pseudo-views are re-projected from the seen views. Through this design, our model is expected to put more trust in the warped pseudo-views that are visible in seen views.  $\phi$ denotes the uncertain embeddings that model per-image uncertain colors, thus each view has its own distinctive $\phi$.

\begin{figure}[t]
  \centering
   \includegraphics[width=0.9\linewidth]{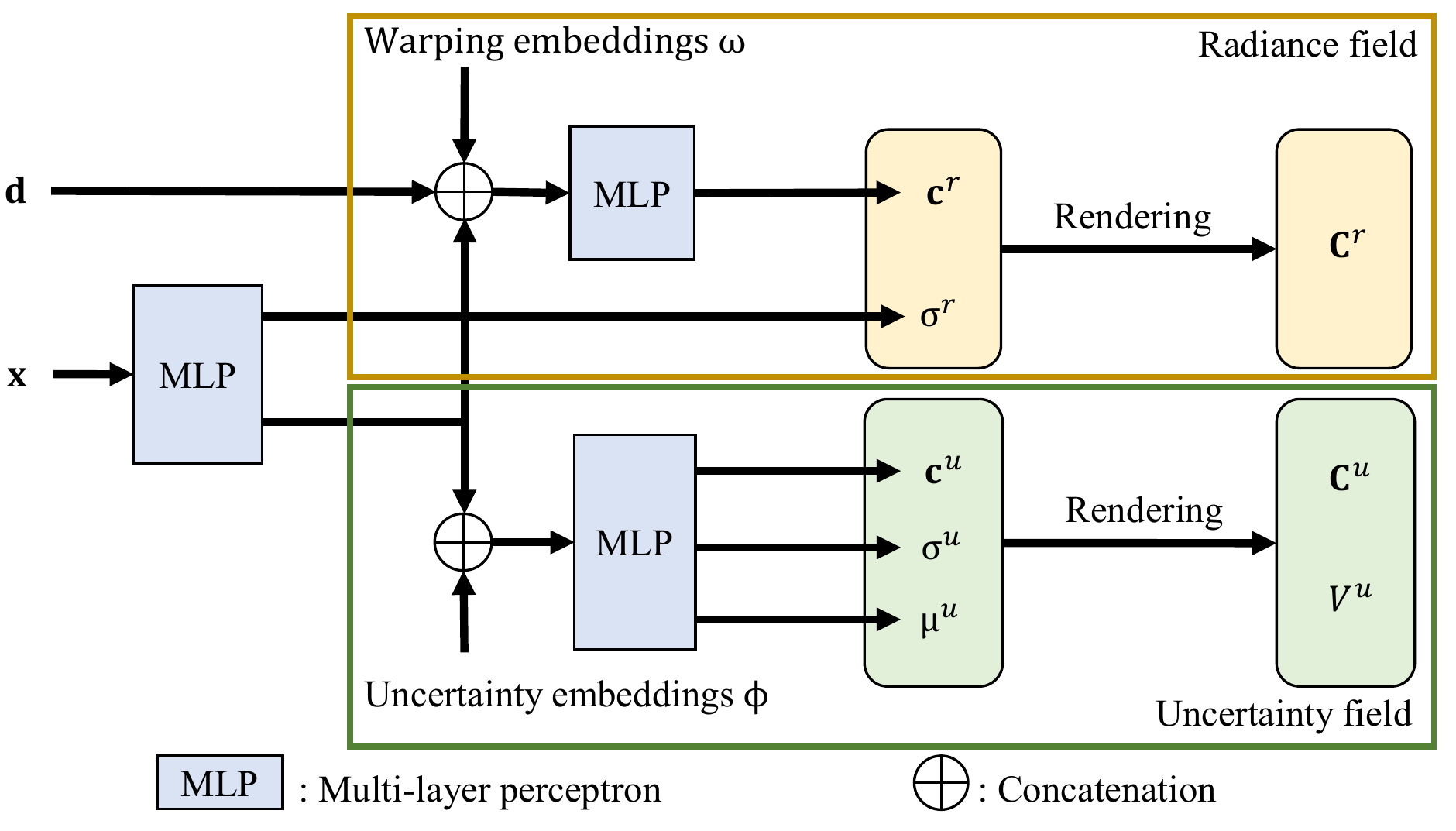}
   \caption{The architecture of our uncertainty-aware model.}
   \label{fig:network}
\end{figure}

Owing to the uncertainty field, our model relaxes mip-NeRF's strict consistency assumption and imposes mip-NeRF to provide larger $\mu^{u}$ in the unobserved regions instead of collapsing to the trivial solution during the iterative process. As a result, Self-NeRF can attenuate the negative impact of uncertainty caused by warping or overfitting and gain information from adequate pseudo-views, yielding a superior image quality. The structure of our model is shown in Fig.~\ref{fig:network}.

\subsection{Inference and optimization}\label{sec:loss} 
\noindent{\textbf{Inference.}} For the query ray $\mathbf{R}$, we model the predicted color with an isotropic normal distribution with mean color $\mathbf{C}^{p}(\mathbf{R})$ and variance $V^{u}(\mathbf{R})$. To get $\mathbf{C}^{p}(\mathbf{R})$ according to Eq.~\ref{eq:sum}, we calculate $\mathbf{C}^{r}(\mathbf{R})$ through Eq.~\ref{eq:static} and the learned variable colors $\mathbf{C}^{u}(\mathbf{R})$ can be analogously rendered with:
\begin{equation}
\label{eq:radient}
\mathbf{C}^{u}(\mathbf{R})=\sum^{N}_{i}\exp(-\sum^{i-1}_{j}\sigma^{u}_j\delta^{u}_j)(1-\exp(\sigma^{u}_i\delta^{u}_i))\mathbf{c}^{u}_i
\end{equation} 
Similar to Eq.~\ref{eq:radient},  $V^{u}(\mathbf{R})$ is approximated with a linear combination of sampled points:
\begin{equation}
V^{u}(\mathbf{R})=\sum^{N}_{i}\exp(-\sum^{i-1}_{j}\sigma^{u}_j\delta^{u}_j)(1-\exp(\sigma^{u}_i\delta^{u}_i))\mu^{u}_i
\end{equation}
In addition, we render the depth $d(\mathbf{R}) \in  \mathbf{D}_i$:
\begin{equation}
d(\mathbf{R})=\sum^{N}_{i}\exp(-\sum^{i-1}_{j}\sigma_j\delta_j)(1-\exp(\sigma_i\delta_i))t_i
\end{equation}
$d(\mathbf{R})$ is leveraged in Eq.~\ref{eq:warp} to generate warped pseudo-views for the subsequent iteration.

\noindent{\textbf{Optimization loss.}} 
For ray $\mathbf{R}$ with the ground truth RGB $\mathbf{C}_{gt}(\mathbf{R})$, the RGB loss is the negative log-likelihood (NLL):
\begin{equation}
\mathcal{L}_{r}=\frac{1}{2}\log (\mu^u)^2+\frac{(\mathbf{C}_{gt}(\mathbf{R}) - \mathbf{C}^{p}(\mathbf{R}))^2}{2(\mu^u)^2}
\end{equation}
Likewise, we obtain the pseudo loss for ray $\mathbf{R}$ with pseudo-views $\mathbf{C}_{pseudo}(\mathbf{R})$ as following:
\begin{equation}
\mathcal{L}_{p}=\frac{1}{2}\log (\mu^u)^2+\frac{(\mathbf{C}_{pseudo}(\mathbf{R}) - \mathbf{C}^{p}(\mathbf{R}))^2}{2(\mu^u)^2} + \frac{\lambda_{u}}{N}\sum_{i}^{N}\sigma^{u}_i
\end{equation}

We further regularize the cone tracing with a cone entropy loss following Shannon Entropy~\cite{Shannon1948AMT}:

\begin{equation}
\mathcal{L}_{c}=-\sum_i^N \frac{1-\exp(\sigma_i\delta_i)}{\sum_j^N 1-\exp(\sigma_j\delta_j)}\log  \frac{1-\exp(\sigma_i\delta_i)}{\sum_j^N 1-\exp(\sigma_j\delta_j)}
\end{equation}

The total loss function to optimize our model is given by:
\begin{equation}
\mathcal{L}=\mathcal{L}_{r}+\lambda_1\mathcal{L}_{p}+\lambda_2\mathcal{L}_{c}
\end{equation}
where $\lambda_1$ and $\lambda_2$ denote manual parameters to balance the loss terms. 
In particular, $\lambda_1$ decays exponentially by a factor of 2 at every 10k steps. The slowly decreasing weight is expected to help the optimization process avoid poor local minima so that pseudo-views provide guidance without conflicting with seen views.

\subsection{Convergence analysis of Self-NeRF}\label{sec:proof}
\begin{figure}[t]
  \centering
   \includegraphics[width=\linewidth]{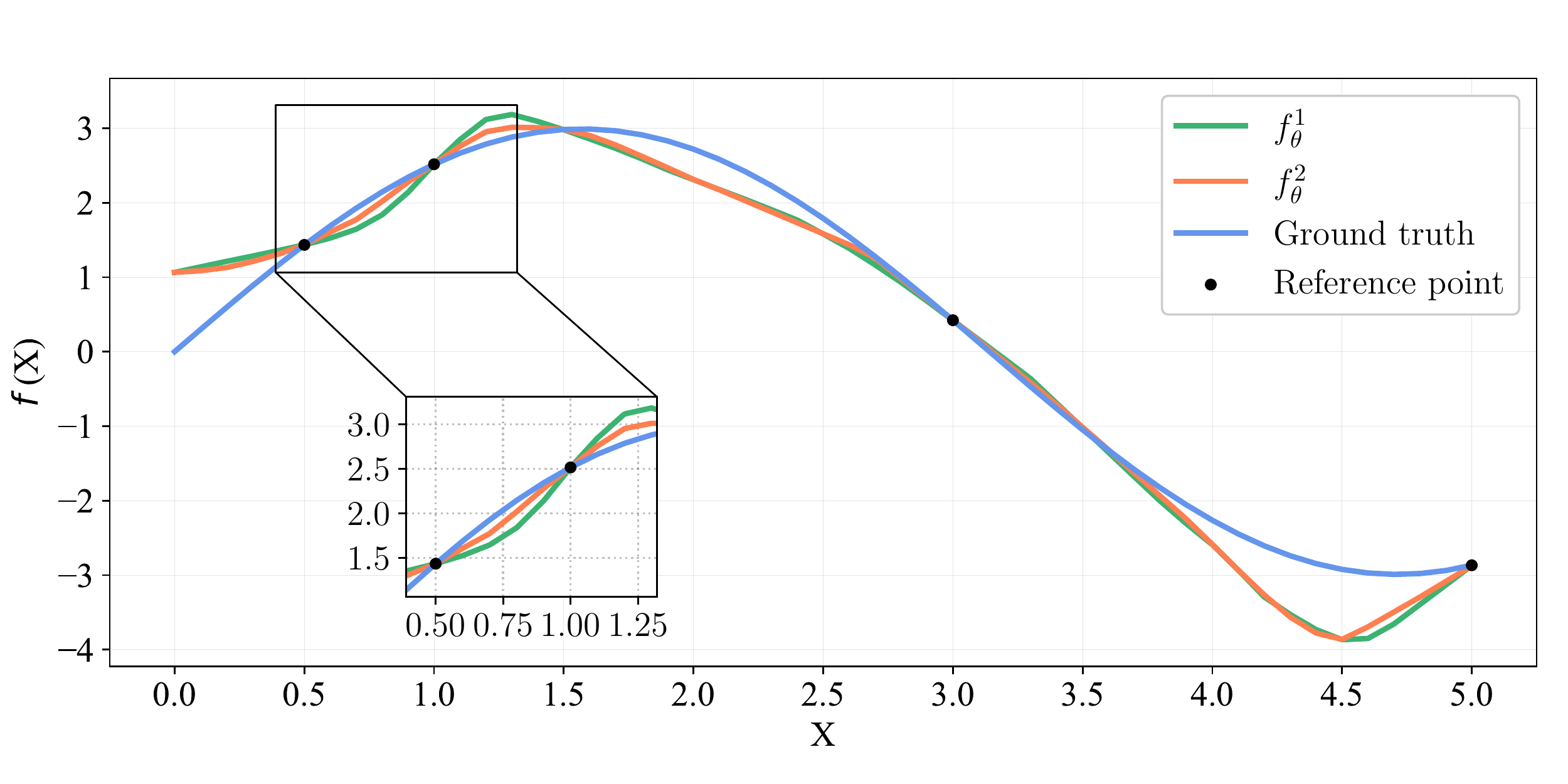}
   \caption{Comparison between the learned models $f_{\theta}^{1}$ and $f_{\theta}^{2}$. We zoom in on the area in the blue box to demonstrate the improvement of $f_{\theta}^{2}$.}
   \label{fig:proof}
\end{figure}
\begin{figure*}[tbp]
  \begin{center}
  \renewcommand\tabcolsep{1.0pt}
  \begin{tabular}{ccccc}
  \includegraphics[width=0.19\linewidth, trim={0px, 0px, 0px, 0px}, clip]{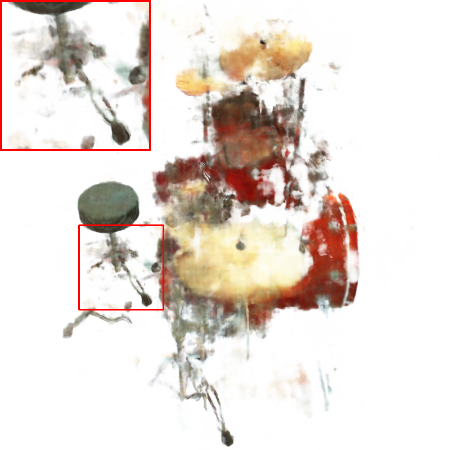}
  &\includegraphics[width=0.19\linewidth, trim={0px, 0px, 0px, 0px}, clip]{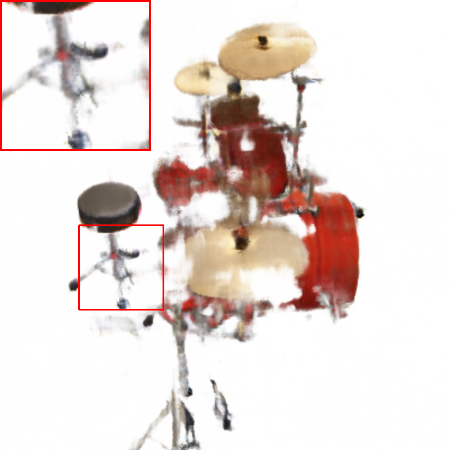}
  &\includegraphics[width=0.19\linewidth, trim={0px, 0px, 0px, 0px}, clip]{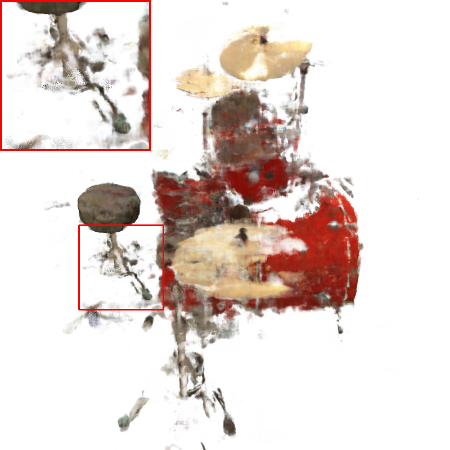}
  &\includegraphics[width=0.19\linewidth, trim={0px, 0px, 0px, 0px}, clip]{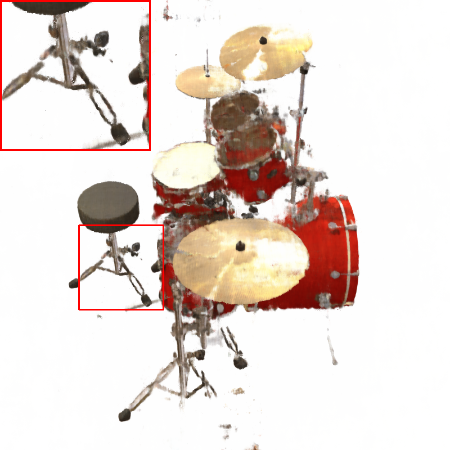}
  &\includegraphics[width=0.19\linewidth, trim={0px, 0px, 0px, 0px}, clip]{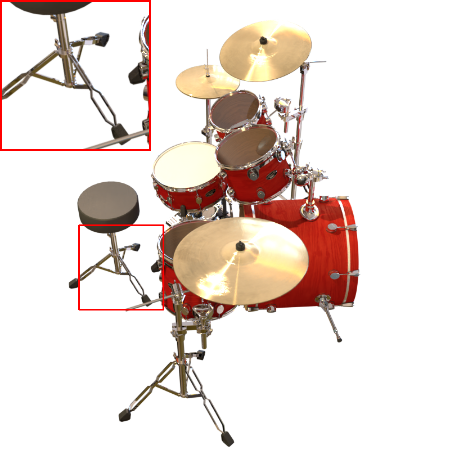}\\
  \includegraphics[width=0.19\linewidth, trim={0px, 0px, 0px, 0px}, clip]{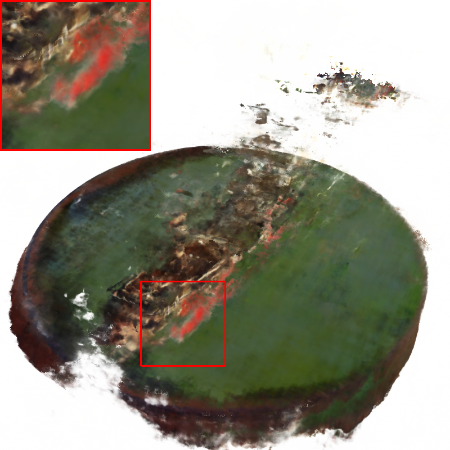}
  &\includegraphics[width=0.19\linewidth, trim={0px, 0px, 0px, 0px}, clip]{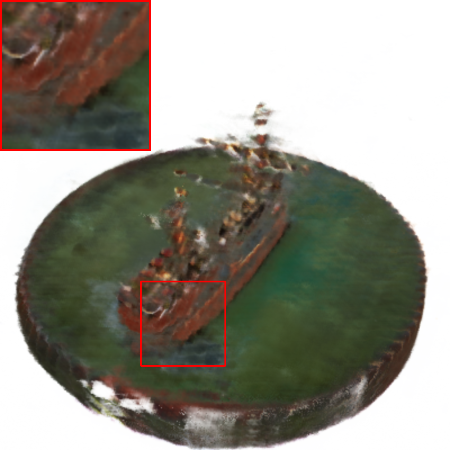}
  &\includegraphics[width=0.19\linewidth, trim={0px, 0px, 0px, 0px}, clip]{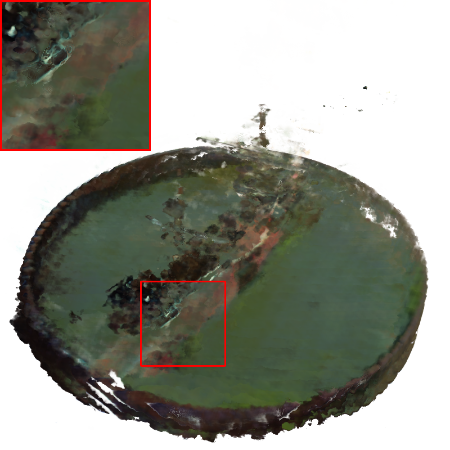}
  &\includegraphics[width=0.19\linewidth, trim={0px, 0px, 0px, 0px}, clip]{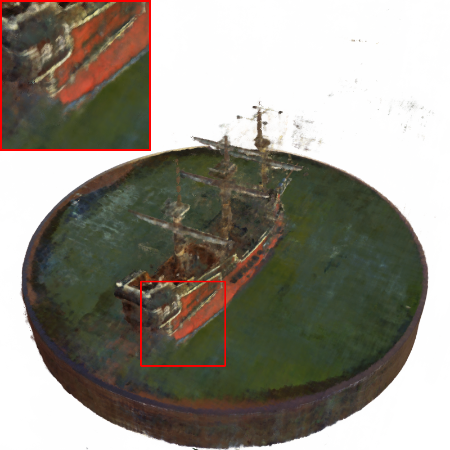}
  &\includegraphics[width=0.19\linewidth, trim={0px, 0px, 0px, 0px}, clip]{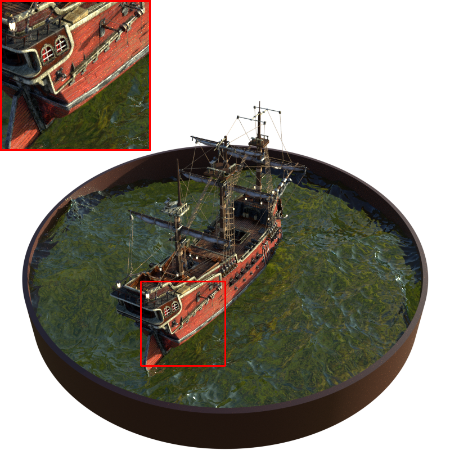}\\
  \includegraphics[width=0.19\linewidth, trim={0px, 0px, 0px, 0px}, clip]{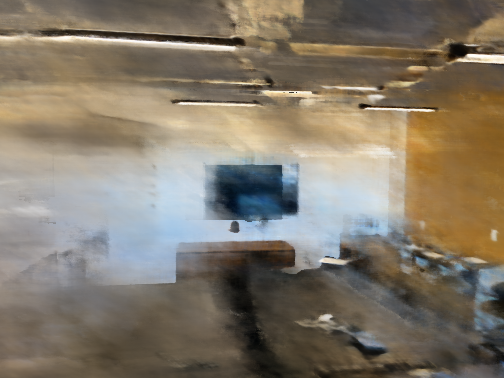}
  &\includegraphics[width=0.19\linewidth, trim={0px, 0px, 0px, 0px}, clip]{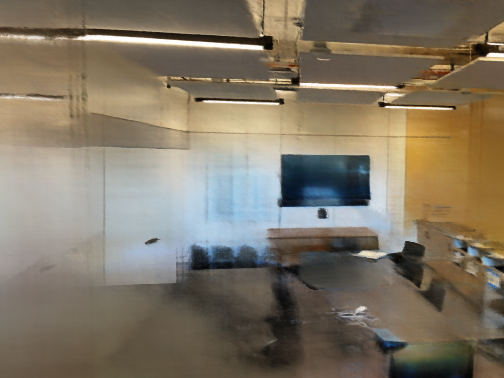}
  &\includegraphics[width=0.19\linewidth, trim={0px, 0px, 0px, 0px}, clip]{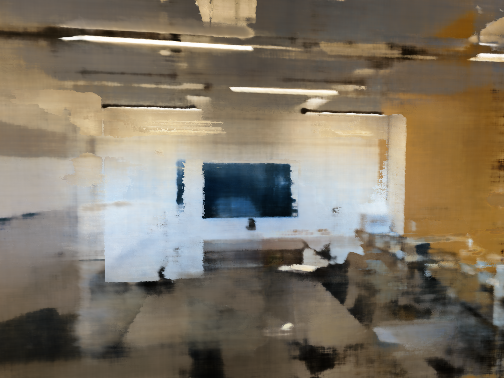}
  &\includegraphics[width=0.19\linewidth, trim={0px, 0px, 0px, 0px}, clip]{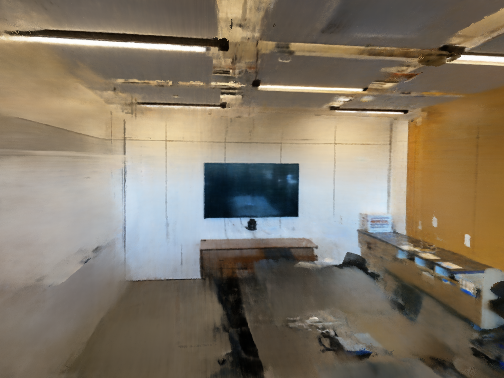}
  &\includegraphics[width=0.19\linewidth, trim={0px, 0px, 0px, 0px}, clip]{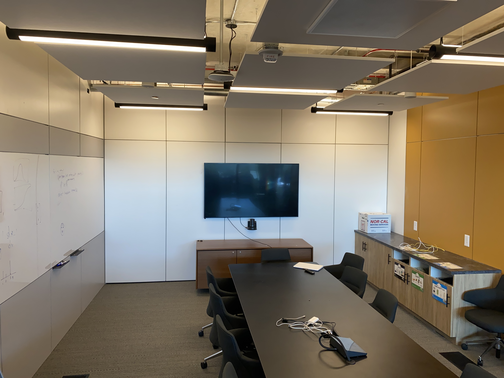}\\
  \includegraphics[width=0.19\linewidth, trim={0px, 0px, 0px, 0px}, clip]{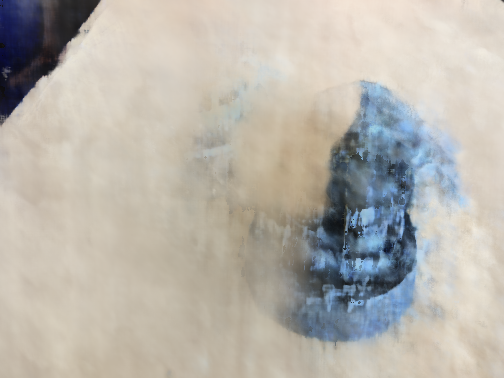}
  &\includegraphics[width=0.19\linewidth, trim={0px, 0px, 0px, 0px}, clip]{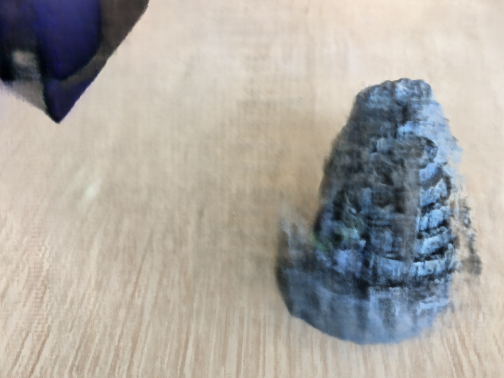}
  &\includegraphics[width=0.19\linewidth, trim={0px, 0px, 0px, 0px}, clip]{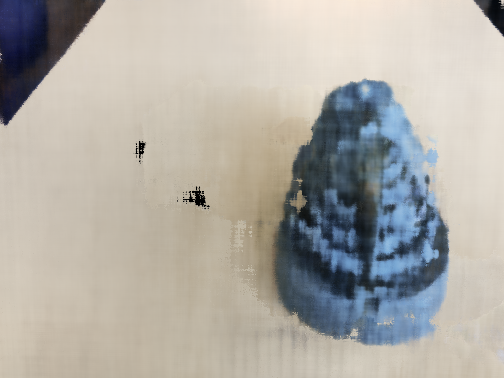}
  &\includegraphics[width=0.19\linewidth, trim={0px, 0px, 0px, 0px}, clip]{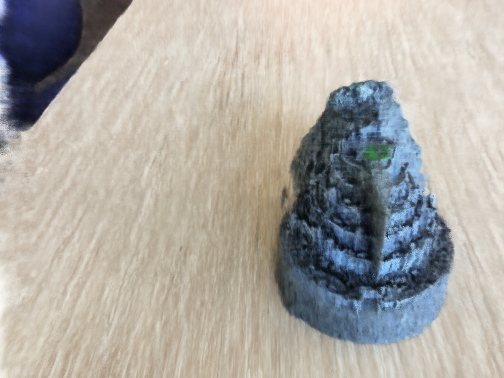}
  &\includegraphics[width=0.19\linewidth, trim={0px, 0px, 0px, 0px}, clip]{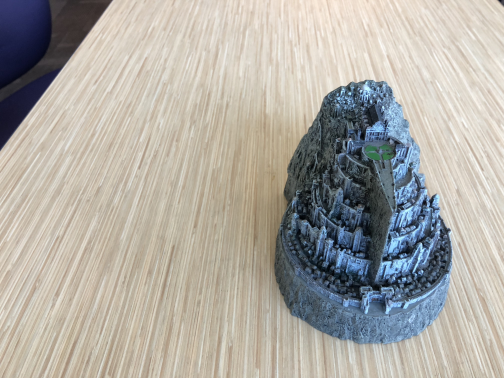}\\
  NeRF~\cite{Mildenhall2020NeRFRS} & DietNeRF~\cite{Jain2021PuttingDiet} & InfoNeRF~\cite{InfoNeRF} & Ours & Ground truth\\
  \end{tabular}
  \end{center}
  
  \caption{Qualitative comparison on the NeRF synthetic dataset~\cite{Mildenhall2020NeRFRS} (Row 1-2) in 4-view settings and LLFF dataset~\cite{Mildenhall2020NeRFRS} (Row 3-4) in 2-view settings. }
  \label{fig:results}
\end{figure*}
Lee \emph{et al.}~\cite{pseudoLabel} have analyzed self-learning technique and proved that it is an effect equivalent to a version of the entropy regularization. The unlabeled data can improve generalization performance even pseudo labels are not precise. For the novel view synthesis task, our pseudo-views reduce overfitting theoretically by providing the possible solution for unseen views. 
In other words, the statement $P(n)$ that $f_\theta^{n}$ outperforms $f_\theta^{(n-1)}$ is true for any $n \in [2,\infty]$. We prove it by induction as follows.

\textbf{Basic step.} We first prove that the statement $P(n)$ is true for $n = 2$. For the sake of simplicity, let us assume that our target function is $F_{gt}(x)=\sin(x-0.1)+\sin(x)+\sin(x+0.1)$, which is an analog of integrating along the rays.
Taking 4 pairs of labeled data as few-shot input, we tend to model $F_{gt}(x)$ with a naive network $f_{\theta}$ through our self-learning pipeline. $f_{\theta}$ comprise three fully-connected layers and has 100 neurons in the hidden layer. Following Sec.~\ref{sec:label}, we gather predicted pseudo-labels $f_\theta^{(i-1)}$ and assign them to unlabeled data $x_u$. 
Apart from that, we mimic warped pseudo-labels using $\frac{1}{2}(f_\theta^{(i-1)}(x_u)+F(x_u))$. Fig.~\ref{fig:proof} shows the performance of learned models $f_\theta^{(1)}$ and  $f_\theta^{(2)}$. $f_\theta^{(1)}$ refers to the model trained solely with few-shot input. Besides the few-shot input, $f_\theta^{(2)}$ further uses 8 warped-labels and 12 predicted pseudo-labels. Noted that $f_\theta^{(2)}$ with a smooth curve is closer to $F_{gt}(x)$ than $f_\theta^{(1)}$.  From a quantitative perspective, $f_\theta^{(2)}$ has lower mean absolute error than $f_\theta^{(1)}$. Therefore, $P(2)$ is proved to be true. 

\textbf{Inductive step.} If the statement $P(k)$ is true where $k\in[2, \infty]$, the prediction of $f_\theta^{k}$ surpasses that of $f_\theta^{(k-1)}$. Hence the pseudo-labels from $f_\theta^{k}$ have a higher signal-noise ratio. When we train $f_\theta^{(k+1)}$ under the same setting as that in training $f_\theta^{(k)}$, the higher quality of pseudo labels ideally leads to a better model. Consequently, $f_\theta^{(k+1)}$ outperforms $f_\theta^{k}$. In other words, the truth of $P(k)$ implies the truth of $P(k+1)$. Therefore, $P(n)$ is true for any $n\in[2,\infty] $. That is to say, the learned model will be improved with our iterative self-training, yielding photo-realistic novel-view synthesis results without additional priors.

Since we do not add priors during iterative training, there exists the upper bound of Self-NeRF. To determine whether Self-NeRF has reached the upper bound, we sample some unseen views as validation dataset. We think that Self-NeRF has converged when the performance of the unseen views does no improve. 

\section{Experiments}
\subsection{Experimental settings}
\noindent{\textbf{Baseline.}}
We compare our method with baseline NeRF~\cite{Mildenhall2020NeRFRS} and two state-of-the-art models for few-shot NeRF, including DietNeRF~\cite{Jain2021PuttingDiet} and InfoNeRF~\cite{InfoNeRF}.

\noindent{\textbf{Dataset.}}
We demonstrate our approach on NeRF synthetic dataset~\cite{Mildenhall2020NeRFRS} and LLFF dataset ~\cite{Mildenhall2019LocalLF}.
In NeRF synthetic dataset, we randomly sample 4 viewpoints as few-shot input for each scene. We run this experiment three times and compute the average scores on 200 testing images for evaluation. For LLFF dataset, we take one out of every eight images from the collection of images for evaluation and randomly select 2 views from the remaining images as training images.

\noindent{\textbf{Evaluation Metrics.}}
We measure the rendered image quality with several quantitative metrics, including the peak signal-to-noise ratio (PSNR), structural similarity index (SSIM) and learned perceptual image patch similarity (LPIPS). PSNR and SSIM are popular metrics for evaluating reconstruction quality, while LPIPS can reflect the perception of humans more precisely.

\noindent{\textbf{Implement details.} }
We implement Self-NeRF with PyTorch~\cite{pytorch} while other approaches run on their own official codes. The Adam optimizer~\cite{Kingma2014AdamAM} is used with an initial learning rate of 0.0005 for optimization. We train other methods for 50K steps (about 15 hours) with their default settings and ensure they have converged. For a fair comparison, we train Self-NeRF in 2 iterations within 15 hours. All these experiments are conducted on a Tesla V100 GPU.

\subsection{Comparisons}

\begin{table}[t]
  \begin{center}
    \caption{Quantitative evaluation of our method against NeRF~\cite{Mildenhall2020NeRFRS}, DietNeRF~\cite{Jain2021PuttingDiet} and InfoNeRF~\cite{InfoNeRF} on the NeRF synthetic dataset~\cite{Mildenhall2020NeRFRS} and LLFF dataset~\cite{Mildenhall2019LocalLF}.}
    \label{tab:results}
    \begin{tabular}{c|c|c|c|c|c} 
     Dataset & Metric & ~\cite{Mildenhall2020NeRFRS} & ~\cite{Jain2021PuttingDiet} & ~\cite{InfoNeRF} &Ours  \\ \toprule
     \multirow{3}{*}{Synthetic}& PSNR $\uparrow$& 16.08 & 15.89 & 18.59 &\textbf{20.66}   \\
     & SSIM $\uparrow$ & 0.795& 0.719 & 0.810 & \textbf{0.840}  \\
     & LPIPS $\downarrow$ & 0.284 & 0.330 & 0.225 &\textbf{0.179}  \\ \midrule
     \multirow{3}{*}{LLFF}& PSNR $\uparrow$& 13.47 & 12.62 & 14.38 & \textbf{15.22}  \\
     & SSIM $\uparrow$ &0.280 & 0.253 & 0.319 & \textbf{0.372}  \\
     & LPIPS $\downarrow$ & 0.607& 0.574 & 0.559 & \textbf{0.443}  
     
    \end{tabular}
  \end{center}
\end{table}

\noindent{\textbf{Quantitative comparisons.}}
Tab.~\ref{tab:results} shows quantitative comparisons of our approach against NeRF, DietNeRF~\cite{Jain2021PuttingDiet} and InfoNeRF~\cite{InfoNeRF}. Please refer to the supplemental materials for the detailed experimental results from individual scenes. 
Due to uncertainty, the blurry renderings produced by NeRF can outperform the visually appealing but incorrect renderings of DietNeRF on average error metrics like PSNR. However, the quantitative results of DietNeRF are still comparable to those of NeRF. InfoNeRF significantly reduces artifacts in the rendering, resulting in improved quantitative results.
Self-NeRF outperforms all other methods in comparison in terms of all the evaluation metrics.

\noindent{\textbf{Qualitative comparisons. }}
Fig.~\ref{fig:results} depicts the rendering images synthesized by different methods.
Compared to other methods, Self-NeRF achieves more realistic rendering in the novel views. 
Specifically, NeRF struggles to accurately reconstruct the scene and often produces blurry and cloudy artifacts. DietNeRF attempts to improve upon this by incorporating priors into the model, resulting in more reasonable and appealing renderings in some cases, such as the front of a ship. However, their use of low-dimensional CLIP~\cite{Radford2021LearningTV} embeddings hinders the model's ability to learn high-frequency details. InfoNeRF yields better results with fewer artifacts by imposing the sparsity on the scene. Despite this, their renderings of novel views still exhibit flaws and lack clear details, which makes the results look unrealistic at first glance. 
By contrast, Self-NeRF preserves the best geometry while generating realistic details. For example, our method successfully reconstructs the front of the ship and the texture of the table.

\subsection{Ablation study}
We validate our design choices by performing an ablation study on two scenes from NeRF synthetic dataset.

\begin{table}[t]
  \begin{center}
    \caption{Quantitative comparison with different models. Here "Base" represents the model in the previous iteration. On this basis, we replace our model with mip-NeRF~\cite{mipnerf} and NeRF-W~\cite{NeRFW} for the current iteration. }
    \label{tab:model_varient}
    \begin{tabular}{c|c|c|c|c|c} 
     Scene&Metric & Base & \cite{mipnerf} & ~\cite{NeRFW} & Ours  \\ \toprule
     \multirow{3}{*}{Mic}& PSNR $\uparrow$& 19.38 & 21.42 & 21.40 &\textbf{23.75}   \\
     & SSIM $\uparrow$ & 0.895& 0.916 & 0.896 & \textbf{0.931}  \\
     & LPIPS $\downarrow$ & 0.180 & 0.139 & 0.171 &\textbf{0.101}  \\ \midrule
     \multirow{3}{*}{Ship}& PSNR $\uparrow$& 19.96 & 19.36 & 17.64 & \textbf{21.26}  \\
     & SSIM $\uparrow$ &0.719 & 0.711 & 0.719 & \textbf{0.757}  \\
     & LPIPS $\downarrow$ & 0.318& 0.364 & 0.388 & \textbf{0.263} \\ \midrule 
    \end{tabular}
    \label{tab:ab_model}
  \end{center}
\end{table}

\begin{figure}[t]
  \begin{center}
  \renewcommand\tabcolsep{1.0pt}
  \begin{tabular}{cccc}
  \includegraphics[width=0.24\linewidth, trim={0px, 0px, 0px, 0px}, clip]{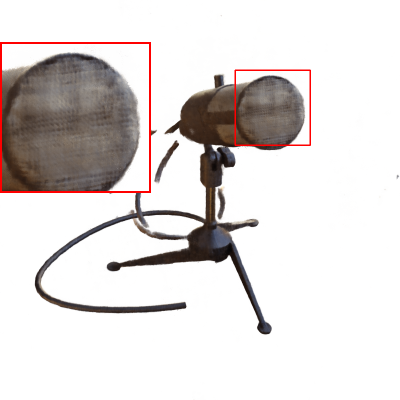}
  &\includegraphics[width=0.24\linewidth, trim={0px, 0px, 0px, 0px}, clip]{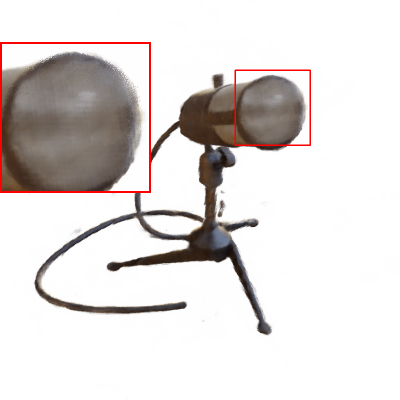}
  &\includegraphics[width=0.24\linewidth, trim={0px, 0px, 0px, 0px}, clip]{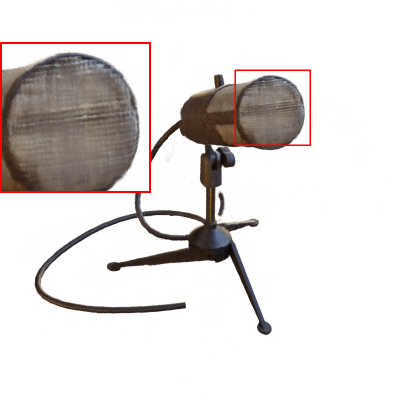}
  &\includegraphics[width=0.24\linewidth, trim={0px, 0px, 0px, 0px}, clip]{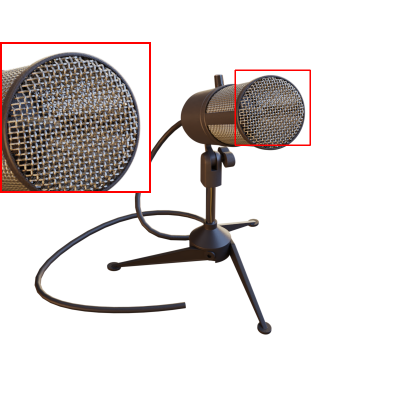}\\
  \small{Mip-NeRF~\cite{mipnerf}} & \small{NeRF-W~\cite{Jain2021PuttingDiet}} & \small{Ours} & \small{Ground truth}\\
  \end{tabular}
  \end{center}
  \caption{Validate the effectiveness of our uncertainty-aware NeRF in Self-NeRF.}
  \label{fig:ab_model}
\end{figure}
\noindent{\textbf{Design for networks. }}
We study the effectiveness of our uncertainty-aware NeRF by replacing it with mip-NeRF~\cite{mipnerf} and NeRF-W~\cite{NeRFW}. Quantitative and qualitative results are given in Tab.~\ref{tab:ab_model} and Fig.~\ref{fig:ab_model}, respectively. We observe that mip-NeRF utilizes cone tracing to capture fine details. However, the absence of uncertainty fields destabilizes the mip-NeRF, resulting in divergent behaviours such as the wire in the mic scene. On the contrary, NeRF-W is an uncertainty-aware model without cone tracing. Thus it alleviates the degradation due to uncertainty but produces blur contents. 
Our model combines the advantages of both and achieves favorable overall performance.

\begin{figure}[t]
  \begin{center}
  \renewcommand\tabcolsep{1.0pt}
  \begin{tabular}{cccc}
  \includegraphics[width=0.24\linewidth, trim={0px, 0px, 0px, 0px}, clip]{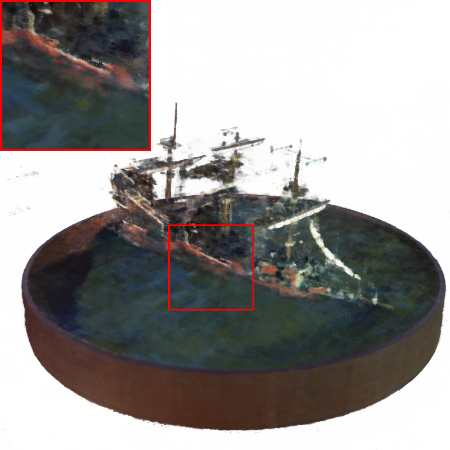}
  &\includegraphics[width=0.24\linewidth, trim={0px, 0px, 0px, 0px}, clip]{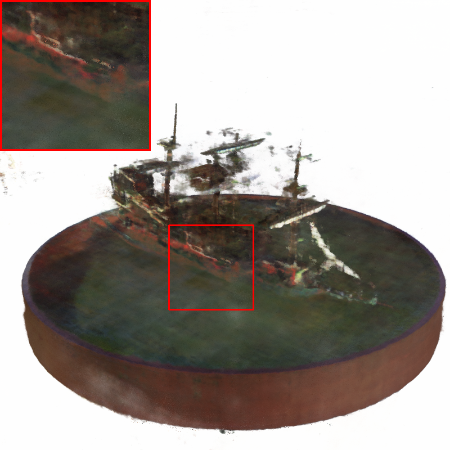}
  &\includegraphics[width=0.24\linewidth, trim={0px, 0px, 0px, 0px}, clip]{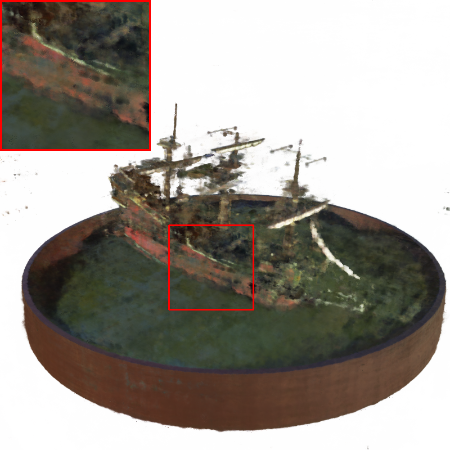}
  &\includegraphics[width=0.24\linewidth, trim={0px, 0px, 0px, 0px}, clip]{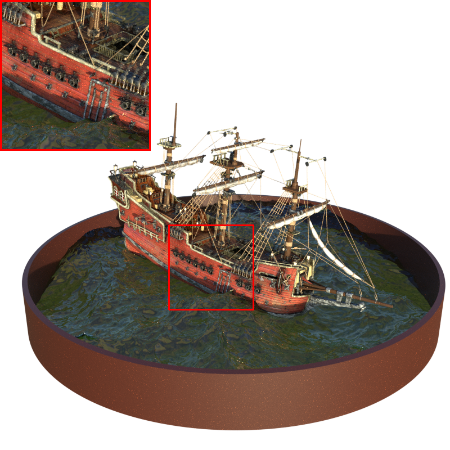}\\
  \small{Predicted} & \small{Warped} & \small{Ours} & \small{Ground truth}\\
  \end{tabular}
  \end{center}
  \caption{Validate the influence of different categories of pseudo-views.}
  \label{fig:ab_label}
\end{figure}

\noindent{\textbf{Choice of pseudo-views. }}
We conduct an ablation study on different categories of pseudo-views. We report quantitative results in Tab.~\ref{tab:ab_label} and show qualitative results in Fig.~\ref{fig:ab_label}. As discussed in Sec.~\ref{sec:label}, the predicted pseudo-views are incapable of accurately reconstructing colors for certain regions, resulting in color deviations when exclusively training with them. In contrast, pixels in the warped pseudo-views are more reliable yet may lack fine-grained details. From the results, it is evident that training with these pseudo-views simultaneously leads to optimal results.

\begin{table}[t]
  \begin{center}
    \caption{Ablation study on different choices of the pseudo-labels. Here "Base" refers to the basic model in the previous iteration. "Warped" refers to training solely with the warped pseudo-views from the basic model while "Predicted" refers to training with the predicted pseudo-views.}
    \label{tab:label_varient}
    \begin{tabular}{c|c|c|c|c|c} 
     Scene&Metric & Base & Warped & Predicted & Ours  \\ \toprule
     \multirow{3}{*}{Mic}& PSNR $\uparrow$& 19.38 & 21.61 & 23.47 &\textbf{23.75}   \\
     & SSIM $\uparrow$ & 0.895& 0.919 & 0.929 & \textbf{0.931}  \\
     & LPIPS $\downarrow$ & 0.180 & 0.128 & 0.103 &\textbf{0.101}  \\ \midrule
     \multirow{3}{*}{Ship}& PSNR $\uparrow$& 19.96 & 20.34 & 20.67 & \textbf{21.26}  \\
     & SSIM $\uparrow$ &0.719 & 0.744 & 0.749 & \textbf{0.757}  \\
     & LPIPS $\downarrow$ & 0.318& 0.290 & 0.270 & \textbf{0.263} \\ \midrule 
    \end{tabular}
    \label{tab:ab_label}
  \end{center}
\end{table}

\subsection{Analysis}

\begin{figure}[t]
  \centering
   \includegraphics[width=1.0\linewidth]{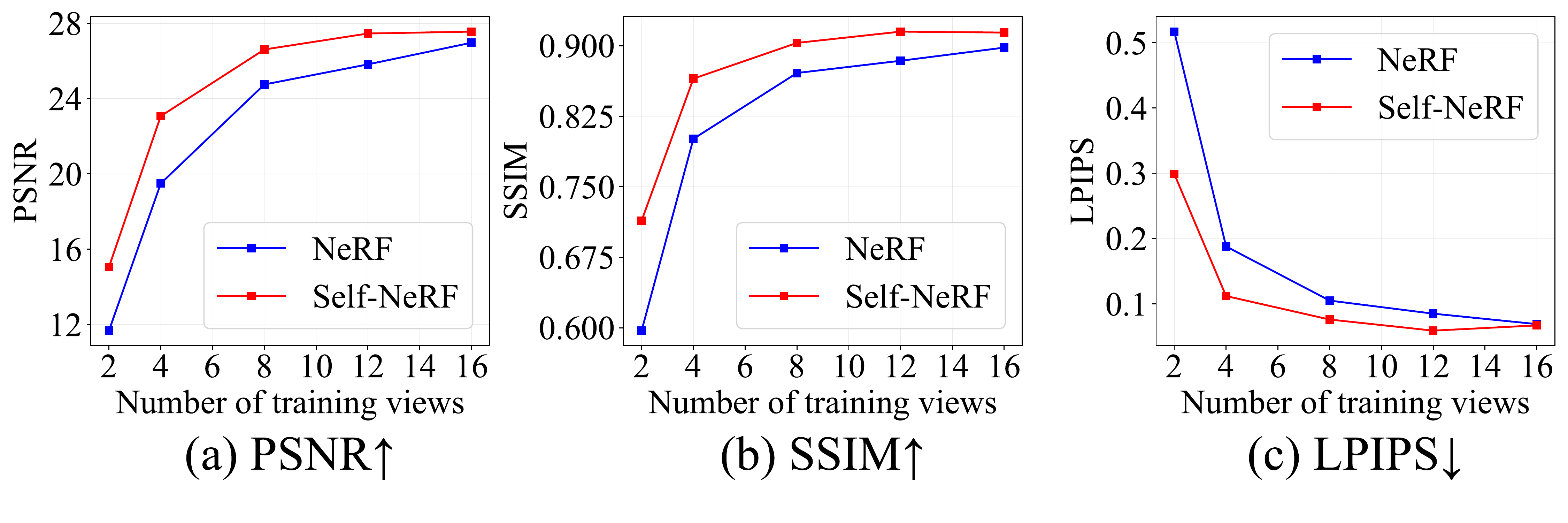}
   \caption{Quantitative results with different numbers of training views.}
   \label{fig:views}
\end{figure}
\noindent{\textbf{Robustness to the number of views. }}
We report the variation curve of quantitative results under different numbers of training views for Lego scene in Fig.~\ref{fig:views}. Our method exhibits gradual performance improvement with an increasing number of training views. While Self-NeRF outperforms NeRF in all metrics, our method's advantage reaches saturation point when using 16 training images. It is partly because the unseen regions decrease as the number of training views increases, thereby limiting the improvement from pseudo-views.

\begin{figure}[t]
  \centering
   \includegraphics[width=1.0\linewidth]{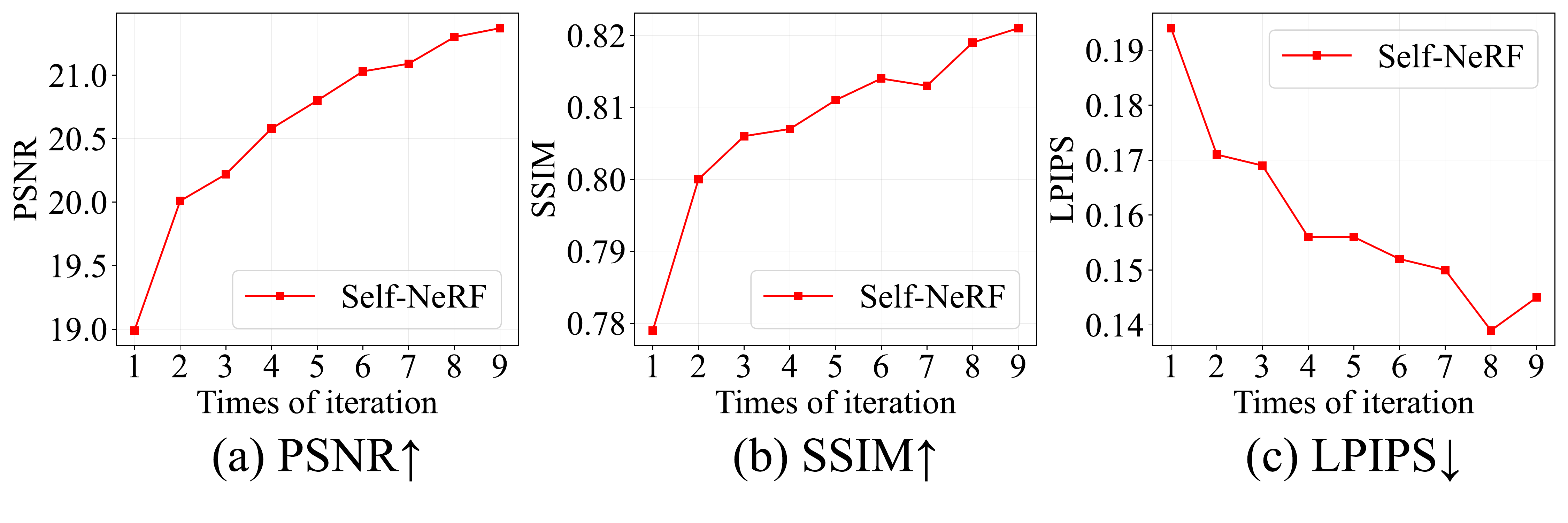}
   \caption{Quantitative results of different iterations.}
   \label{fig:ab_tab_iteration}
\end{figure}
\begin{figure}[tbp]
  \begin{center}
  \renewcommand\tabcolsep{1.0pt}
  \begin{tabular}{cccc}
  \includegraphics[width=0.3\linewidth, trim={0px, 0px, 0px, 0px}, clip]{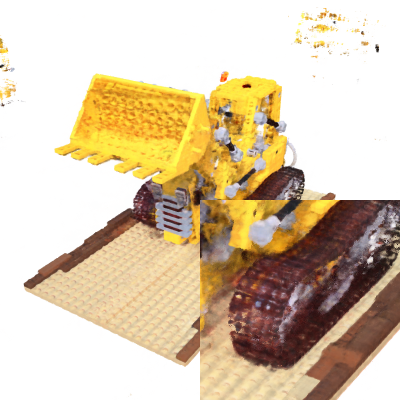}
  &\includegraphics[width=0.3\linewidth, trim={0px, 0px, 0px, 0px}, clip]{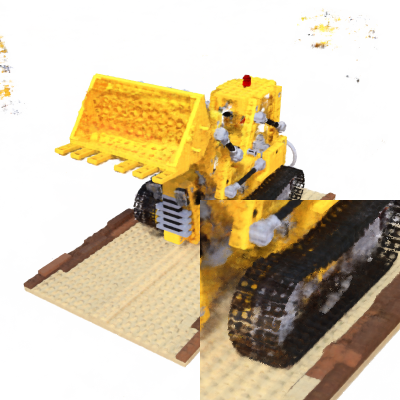}
  &\includegraphics[width=0.3\linewidth, trim={0px, 0px, 0px, 0px}, clip]{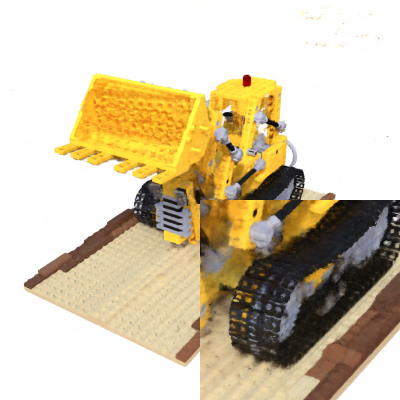}\\
  \includegraphics[width=0.3\linewidth, trim={0px, 0px, 0px, 0px}, clip]{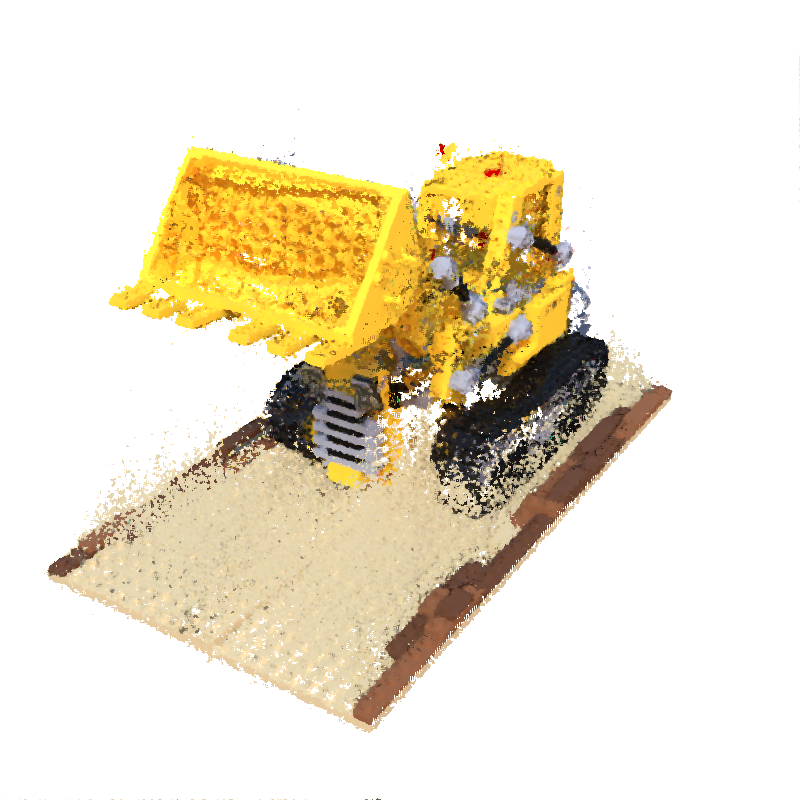}
  &\includegraphics[width=0.3\linewidth, trim={0px, 0px, 0px, 0px}, clip]{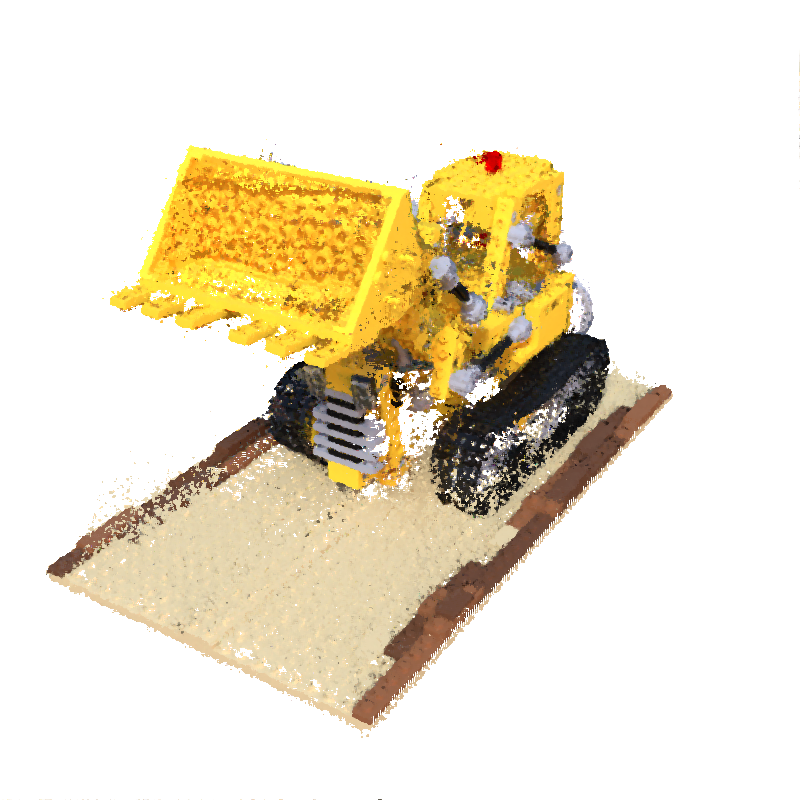}
  &\includegraphics[width=0.3\linewidth, trim={0px, 0px, 0px, 0px}, clip]{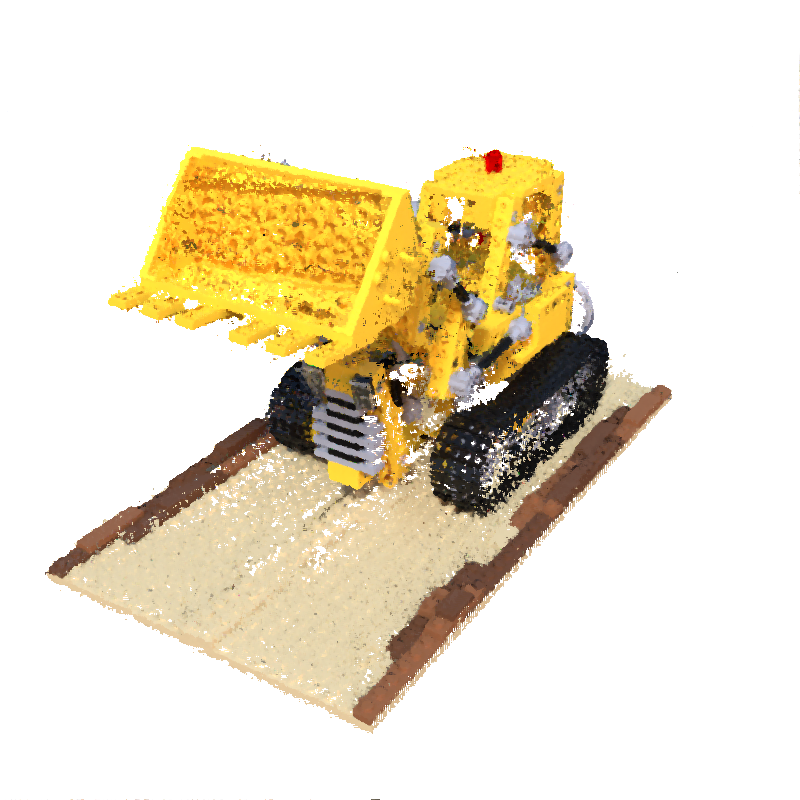}\\
  \includegraphics[width=0.3\linewidth, trim={0px, 0px, 0px, 0px}, clip]{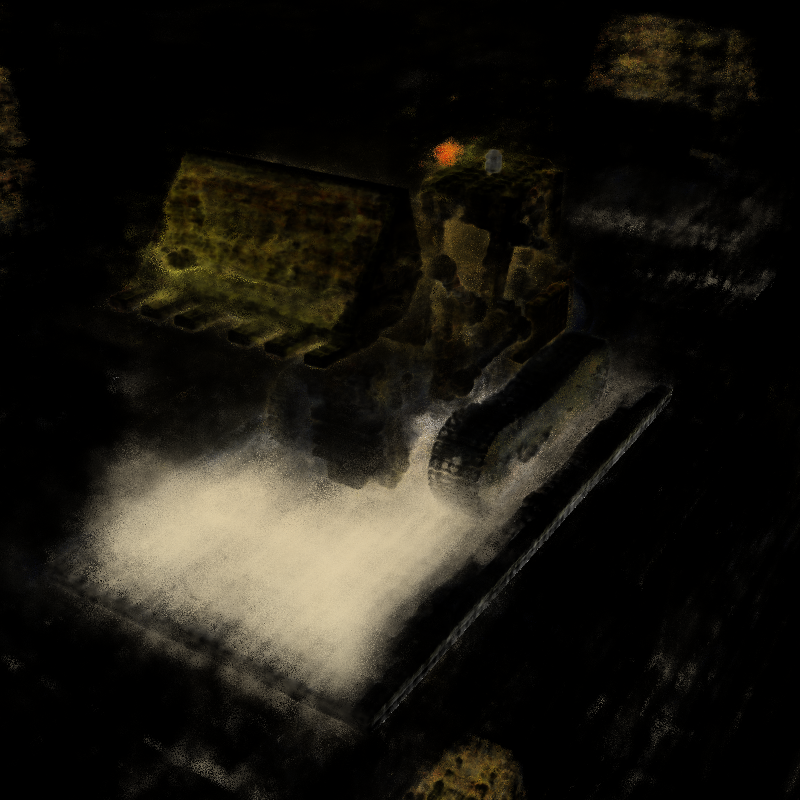}
  &\includegraphics[width=0.3\linewidth, trim={0px, 0px, 0px, 0px}, clip]{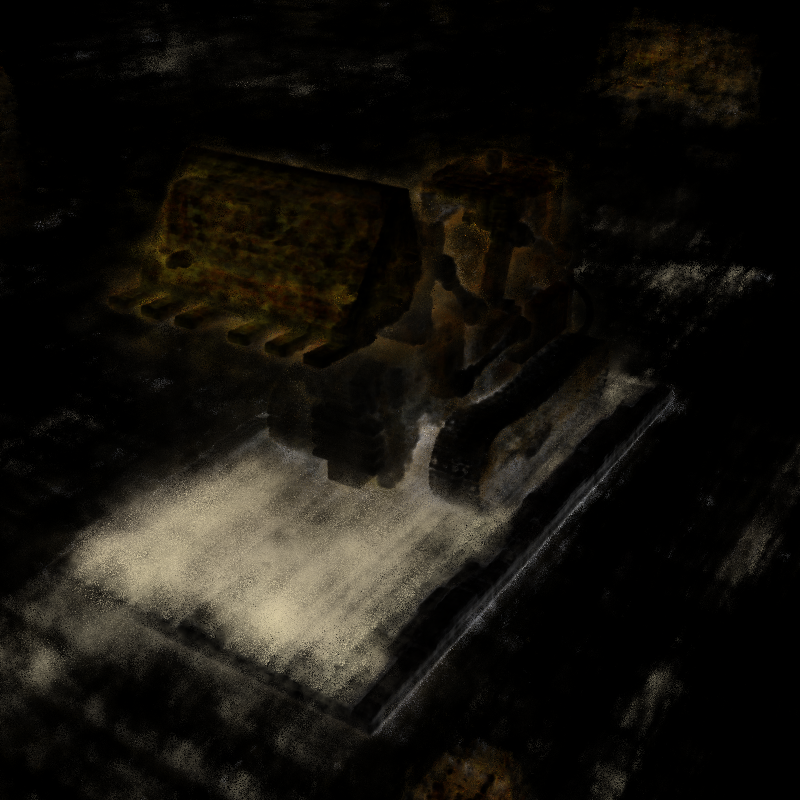}
  &\includegraphics[width=0.3\linewidth, trim={0px, 0px, 0px, 0px}, clip]{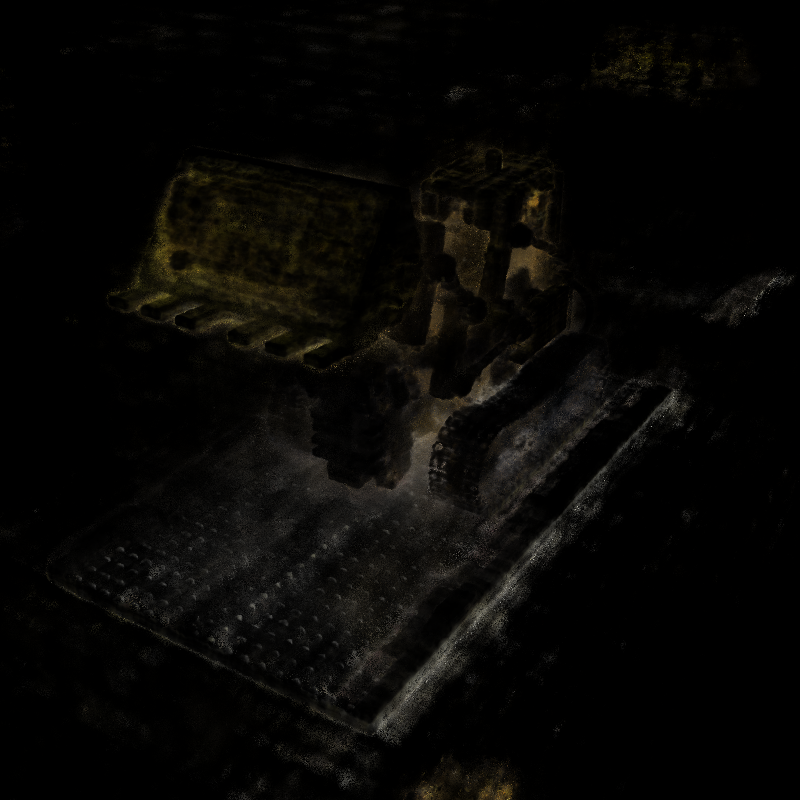}\\
  \small{1 iteration } & \small{2 iterations} & \small{5 iterations} \\
  \end{tabular}
  \end{center}
  \caption{Visualization of the prediction of Self-NeRF (top row), the warped pseudo-views (middle row) and the corresponding uncertainty map (bottom row) in different iterations. With the increasing number of iterations, Self-NeRF corrects the color shifts on the wheel and reduces artifacts. The improved performance results in better pseudo-views with less uncertainty, which in turn benefits the training.}
  \label{fig:ab_iteration}
\end{figure}
\noindent{\textbf{Improvement in iterative training. }}
We report the quantitative results of various iterations in Fig.~\ref{fig:ab_tab_iteration} and depict the outputs of Self-NeRF in Fig.~\ref{fig:ab_iteration}. 
Self-NeRF has converged since the LPIPS deteriorates in the $9^{th}$ iteration.
Note that our uncertainty-aware NeRF is capable of detecting uncertain pixels and leverages pseudo-views to their full potential. It gradually reduces the artifacts and effectively mitigates color shifts as the number of iterations increases. Hence, our iterative process leads to continuous improvement of the overall quality of predictions.

\section{Conclusion}
In this paper, we propose Self-NeRF to synthesize novel views given few-shot images. Inspired by self-training, Self-NeRF iteratively generates pseudo-views and trains the model with seen views and pseudo-views jointly.
In the iteration, we generate two categories of pseudo-views: predicted pseudo-views from the previous iteration and warped pseudo-views which are reprojected from seen views using depth-based forward warping. These pseudo-views are shown to have a stabilizing effect and alleviate the color shifts. 
To avoid the negative impact of uncertain pixels in pseudo-views, we propose an uncertainty-aware NeRF with specialized embeddings. We also utilize techniques such as cone entropy regularization to reconstruct fine details and facilitate optimization. 
Our experiments further demonstrate our method’s competitiveness compared with state-of-the-art models for few-shot novel view synthesis.
{\small
\bibliographystyle{ieee_fullname}
\bibliography{egbib}
}

\end{document}